\documentclass[preprint]{elsarticle}
\usepackage{hyperref}

\hypersetup{colorlinks = true,linkcolor = blue,anchorcolor =red,citecolor = blue,filecolor = red,urlcolor = red,
            pdfauthor=author}

\newpageafter{abstract}

\begin{document}

\begin{frontmatter}

\title{Benchmarking AutoML algorithms on a collection of synthetic classification problems }

\author[inst1]{Pedro Henrique Ribeiro}
\ead{pedro.ribeiro@cshs.org}
\author[inst2,inst3]{Patryk Orzechowski}
\ead{patryk.orzechowski@gmail.com}
\author[inst2]{Joost Wagenaar}
\ead{joostw@seas.upenn.edu}
\author[inst1]{Jason H. Moore\corref{cor1}}
\ead{jason.moore@csmc.edu}

\cortext[cor1]{Corresponding author}

\affiliation[inst1]{organization={Department of Computational Biomedicine, Cedars-Sinai Medical Center},
            addressline={700 N San Vicente Blvd, Pacific Design Center}, 
            city={West Hollywood},
            postcode={90069}, 
            state={CA},
            country={USA}}

\affiliation[inst2]{organization={Department of Biostatistics, Epidemiology and Bioinformatics, University of Pennsylvania},
addressline={3700 Hamilton Walk}, 
            city={Philadelphia},
            postcode={19104}, 
            state={PA},
            country={USA}}

\affiliation[inst3]{organization={Department of Automatics and Robotics, AGH University of Science and Technology},
addressline={al. Mickiewicza 30}, 
            city={Krakow},
            postcode={30-059}, 
            country={Poland}}

\begin{abstract}

Automated machine learning (AutoML) algorithms have grown in popularity due to their high performance and flexibility to adapt to different problems and data sets. With the increasing number of AutoML algorithms, deciding which would best suit a given problem becomes increasingly more work. Therefore, it is essential to use complex and challenging benchmarks which would be able to differentiate the AutoML algorithms from each other. This paper compares the performance of four different AutoML algorithms: Tree-based Pipeline Optimization Tool (TPOT), Auto-Sklearn, Auto-Sklearn 2, and H2O AutoML. We use the Diverse and Generative ML benchmark (DIGEN), a diverse set of synthetic datasets derived from generative functions designed to highlight the strengths and weaknesses of the performance of common machine learning algorithms. We confirm that AutoML can identify pipelines that perform well on all included datasets. Most AutoML algorithms performed similarly; however, there were some differences depending on the specific dataset and metric used.

\end{abstract}

\begin{keyword}
Machine Learning \sep AutoML \sep Data Science \sep Pipeline Optimization \sep Artificial Intelligence
\end{keyword}

\end{frontmatter}

\section{Introduction}

Data scientists have a wealth of algorithms at their disposal when designing machine learning pipelines. These include methods for prepossessing, data transformation, feature selection, and, finally, the choice of machine learning algorithms. Each step may also include numerous parameter settings that must be explored and optimized for each dataset. As more tools get developed, the space of possible combinations of algorithms grows. As datasets get larger, the time it takes to explore the optimal combination of these steps also grows. In response, Automated Machine Learning (AutoML) tools have been increasingly viable as new sophisticated search algorithms are developed and computers grow in computational power. Over the years, multiple AutoML algorithms were developed, including TPOT \cite{OlsonGECCO2016}, Auto-Sklearn 1 \cite{feurer-neurips15a}, and Auto-Sklearn 2 \cite{feurer-arxiv20a}, H2O AutoML \cite{H2OAutoML20}.

AutoML aims to create an accessible yet powerful off-the-shelf algorithm to serve as a jump start for a wide array of problems. AutoML aims to produce higher-performing models quicker by more efficiently searching through the space of possible models. These features can increase the accessibility of machine learning analysis to a broader range of people.

When evaluating AutoML tools, it is important to benchmark them against diverse datasets. There have been several benchmarks that compile challenging datasets. Open Machine Learning \cite{OpenML2013} host several benchmarks on their website including AutoML Benchmark by \cite{4ae49c3aade448858065f3a0e1bd3cee}, AMLB by \cite{https://doi.org/10.48550/arxiv.2207.12560}, and AutoML Benchmark Training Datasets from \cite{feurer-arxiv20a}. Other examples include University of California Irvine (UCI) machine learning repository \cite{Dua:2019}, Library for Support Vector Machines (LIBSVM) \cite{chang2011libsvm},Penn Machine Learning Benchmark (PMLB) \cite{olson2017pmlb}, NAS-Bench-360 \cite{DBLP:journals/corr/abs-2110-05668}. There have also been multiple AutoML competitions organized, with the winner being the algorithm that yields the highest performance on a hidden test set, such as the ChaLearn AutoML Challenge Series from 2015-2018 \cite{Guyon2019}. Currently ongoing at the time of writing is also the AutoML Decathlon at the NeurIPS'22 Competition Track \cite{automl-decathlon}.

Noteably, these collections of datasets consist mostly of real data where it is difficult to know the ground truth generative function. Synthetic data where we know the underlying function may allow us to more easily pinpoint where models fail and indetify why. It is also informative to investigate problems with a significant difference in the performance of the tested algorithms. Such problems may lead to insight for developers into potential areas of improvement and, for users, recommendations on which algorithms to use in a given circumstance. This was the motivation behind the DIverse and GENerative ML Benchmark (DIGEN) which provides forty synthetic datasets with diverse generative functions designed to separate the performance of common machine learning algorithms \cite{doi:10.1126/sciadv.abl4747}. DIGEN provides an excellent opportunity to test how robust AutoML learning algorithms are to a diverse set of problems. DIGEN also provides a Python package, found here \url{https://github.com/EpistasisLab/digen}, with code to benchmark machine learning algorithms and produce summary figures. This paper compares the performance of state-of-the-art AutoML algorithms on these datasets.

\section{Materials and Methods}

This study used four popular AutoML algorithms: TPOT, Auto-Sklearn, Auto-Sklearn 2, and H2O AutoML on a new collection of forty synthetic datasets called DIGEN.  

\subsection{AutoML algorithms}

\paragraph{TPOT} Tree-based Pipeline Optimization Tool (TPOT) utilizes evolutionary algorithms to optimize a tree-shaped pipeline of common machine learning steps. It can also be configured with a template to restrict it to a certain shape and sequence of module categories (grouped by selectors, transformers, and classifiers) \cite{OlsonGECCO2016,olson2018system,Olson2016EvoBio,le2020scaling,pmlr-v64-olson_tpot_2016,10.1093/bioinformatics/btz470}. 

We compared Three different configurations of TPOT, each with a different restriction on the pipeline shape:
\begin{enumerate}
    \item \emph{TPOT\_C} - TPOT restricted to only using a single Classifier. That is, no feature selection or prepossessing steps.
    \item \emph{TPOT\_STC} - TPOT with a template consisting of a feature Selector followed by a feature Transformer ending in a  Classifier.
    \item \emph{TPOT\_Base} - The default TPOT with no template and thus can generate unrestricted pipeline shapes.
\end{enumerate}

\paragraph{Auto-Sklearn} Auto-Sklearn uses Bayesian Optimization and meta-learning to identify best-performing machine learning algorithms, perform hyperparameter optimization, and generate ensembles \cite{feurer-neurips15a}. 

\paragraph{Auto-Sklearn 2} Auto-Sklearn 2 builds upon the meta-learning algorithm with Portfolio Successive Halving and automates more decisions with a Policy Section algorithm. These changes led to increased speeds and performance \cite{feurer-arxiv20a}.

\paragraph{H2O AutoML} H2O AutoML uses random search to identify and train base models, which a meta learner then selects from to combine into a stacked ensemble \cite{H2OAutoML20}.

\subsection{Benchmarking data - DIGEN}

Diverse and Generative ML Benchmark (DIGEN) provides 40 mathematical functions that each map ten input features into a binary classification \cite{doi:10.1126/sciadv.abl4747}. These mathematical functions were designed to maximize the diversity of the performance and ranking across several common hyperparameter-tuned machine-learning algorithms. The datasets, therefore, highlight the strengths and weaknesses of the included machine learning algorithms.

DIGEN provides a fixed seed and dataset for each generative function for consistency of comparison across different methods. The seed was also included in the optimization of the dataset. The training data for all datasets consists of 800 samples and ten features. All values are drawn from a normal distribution with no missing values or added noise. For each dataset, the number of informative features ranges from 2 to 9. It is important to note that not all DIGEN datasets are deterministic. When DIGEN generated its original 1000 samples, it determined the labels by sorting the output of the generative function and assigning the bottom and top halves of the values 0 and 1, respectively, to yield an exact 50\% split in target classes. When the smallest value in the upper half and the largest value in the lower half are different, a median threshold exists that allows a function to predict labels perfectly. However, when the median value exists in both the top and bottom halves of the sort, they are effectively randomly assigned to one label or the other. For example, in the digen29\_8322 dataset, the generative function leads to 691 ones and 309 zeros. To even the classes, a random set of 191 ones, 19.1\% of the dataset, gets reclassified to zero. This step could be interpreted as adding noise to the targets. Fourteen datasets in total have similar directional noise. Of those, 13 have less than 5\% of the labels randomly reassigned, and one, digen29\_8322, has 19.1\% of labels reassigned. Both the built-in training and testing data have this noise.

Scoring based on a small dataset is prone to noise or small fluctuations that can lead to larger inaccurate estimates. The original test set built into DIGEN includes only 200 samples. It is not unlikely that a different sample of 200 test points would yield different results. It is also possible for small datasets to favor specific models purely by chance. With a larger dataset, the probability of a biased estimate of the score is reduced. For this study, we took advantage of the synthetic nature of this dataset to regenerate a new test set with 10000 samples and removed the noise to get a more accurate measure of model performance. We calculated the median value as a threshold for binarization from the generation of the original dataset. In the dataset where some of the medians were randomly assigned, we assigned them to the label to which they were most often assigned in the original dataset.

\subsection{Experiment design}

\begin{table}[ht!]
 \scriptsize
\centering
\caption{AutoML Algorithms and their parameters.}
\label{tab:ml-params}
\begin{tabular}{|c|l|p{0.5\linewidth}|}
\hline
{\bf Algorithm} & {\bf Parameters} & {\bf Values}\\
\hline

\textbf{TPOT} &  `template' & \textbf{TPOT\_Base} : None \newline \textbf{TPOT\_C} : `Classifier' \newline
\textbf{TPOT\_STC} : `Selector-Transformer-Classifier\\
  &  `population\_size' & 100 \\
  &  `generations' & 100 \\
  &  `scoring'& `roc\_auc' \\
  &  `n\_jobs'& 48 \\
  &  `verbosity' & 2 \\
  &  `cv' & 10 \\
   & `max\_time\_mins' & 1200/60 \\
  \hline
\textbf{Auto-Sklearn} & 
      `n\_jobs' & 48 \\
      &   `metric' & autosklearn.metrics.roc\_auc \\
      &   `resampling\_strategy\_arguments' & \{`cv': 10\} \\
       &  `time\_left\_for\_this\_task' & 1200  \\
       &  `memory\_limit' & 1000000 \\
      \hline
\textbf{Auto-Sklearn 2} & `n\_jobs' & 48 \\
       &  `metric'& autosklearn.metrics.roc\_auc \\
      &   `time\_left\_for\_this\_task'& 1200 \\
       &  `memory\_limit'& 1000000 \\
              \hline
\textbf{H2O AutoML} & 
    `stopping\_metric' & `AUC' \\
   &  `sort\_metric' & `AUC' \\
   &  `nfolds'& 10 \\
   &  `max\_runtime\_secs' & 1200 \\
   & `max\_mem\_size' (h2o.init() ) & `1000G' \\
   &  `n\_jobs' (h2o.init() ) & 48 \\
    \hline
\end{tabular}
\label{table:params}
\end{table}

Using the $evaluate$ function provided by DIGEN \cite{doi:10.1126/sciadv.abl4747}, each AutoML algorithm was fit to the provided 800-sample training set. We then tested the fitted model on our larger 10000 sample test set. We repeat this ten times for each dataset and model. Parameters were set for each AutoML algorithm to make the experimental setup as similar as possible. All algorithms were set to optimize Area Under the Receiver Operating Curve (AUROC) using 10-fold cross-validation (except for Auto-Sklearn 2, which uses meta-learning to decide the number of folds). All algorithms were set to use 48 threads through the n\_jobs parameter. H2O, Auto-Sklearn 1 and 2 require a memory limit which was set to 1000GB, the full capacity of the compute node. A 20-minute time limit for fitting the model was provided using the appropriate parameter for each model. As some algorithms occasionally went over time, a terminate signal was called after 120\% of the time limit had passed for TPOT and H2O. This signal was not used with Auto-Sklearn as it does not support manual termination via signals. The exact settings of parameters for AutoML algorithms are presented in Table \ref{table:params}.

All models were run on a High Performance Computing cluster node with an Intel Xeon Gold 6342 CPU with 48 threads and 1TB of memory.

\section{Results}

{\noindent
\begin{figure}[]
    \centering
    \includegraphics[scale=0.3]{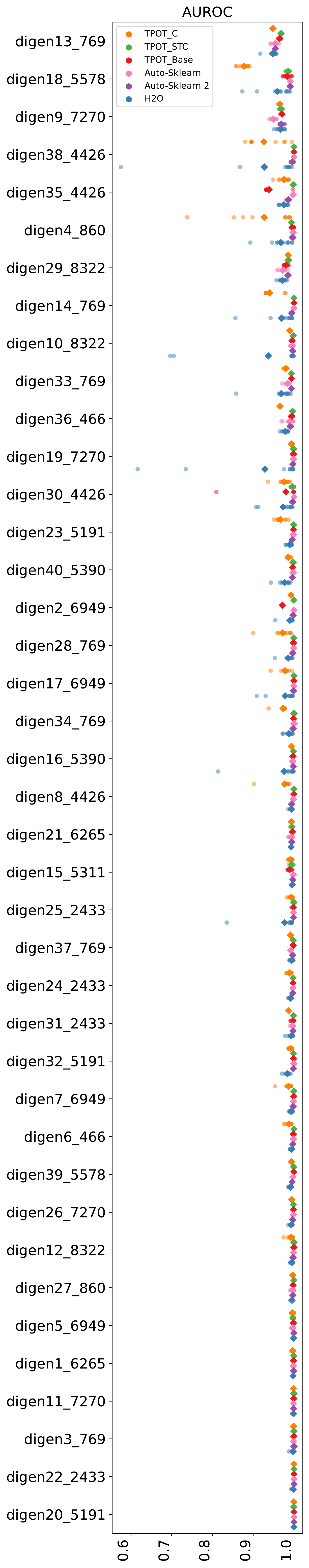}
    \includegraphics[scale=0.3]{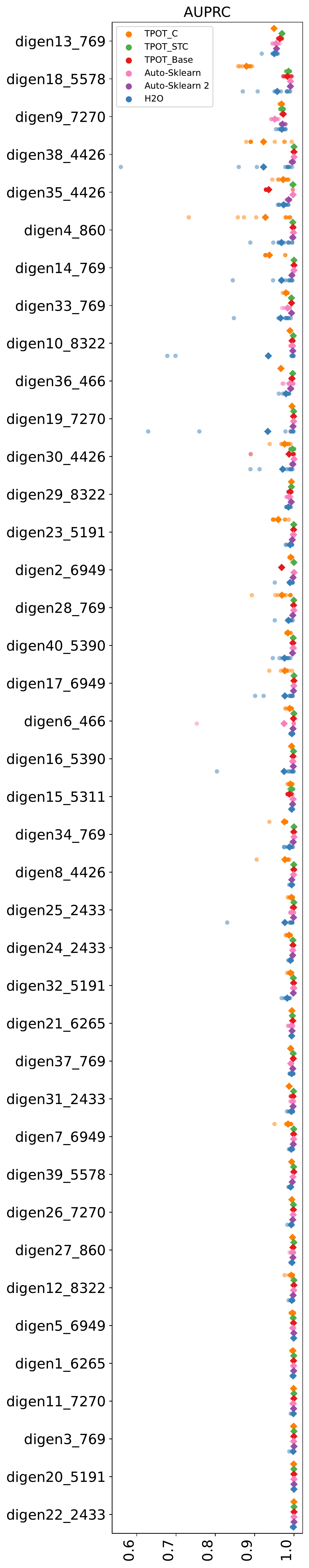}
    \includegraphics[scale=0.3]{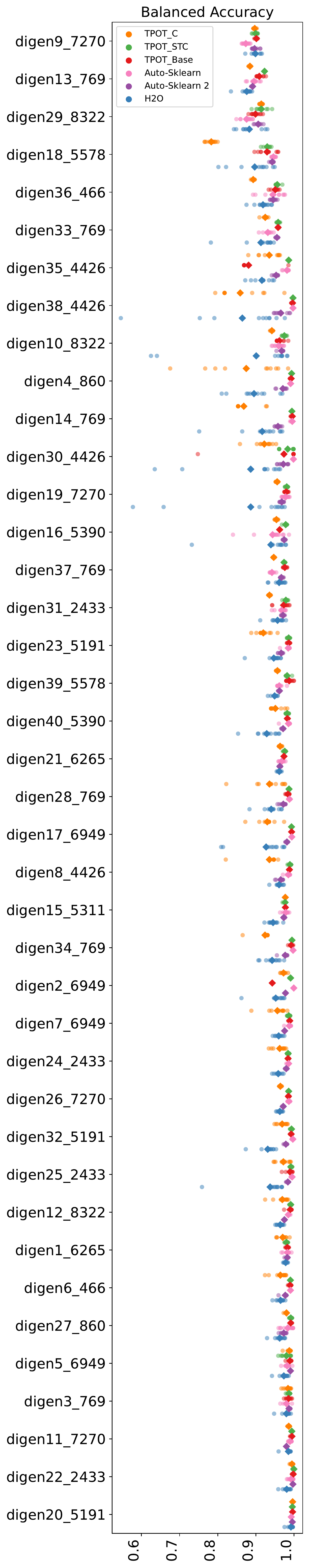}
    \caption{Scores for all runs of each AutoML algorithm and dataset. Diamonds represent the average scores for each AutoML algorithm on each dataset.}
    \label{fig:strip_line}
\end{figure}}

\begin{figure}[htb!]
    \centering
    \centerline{\includegraphics[scale=0.35]{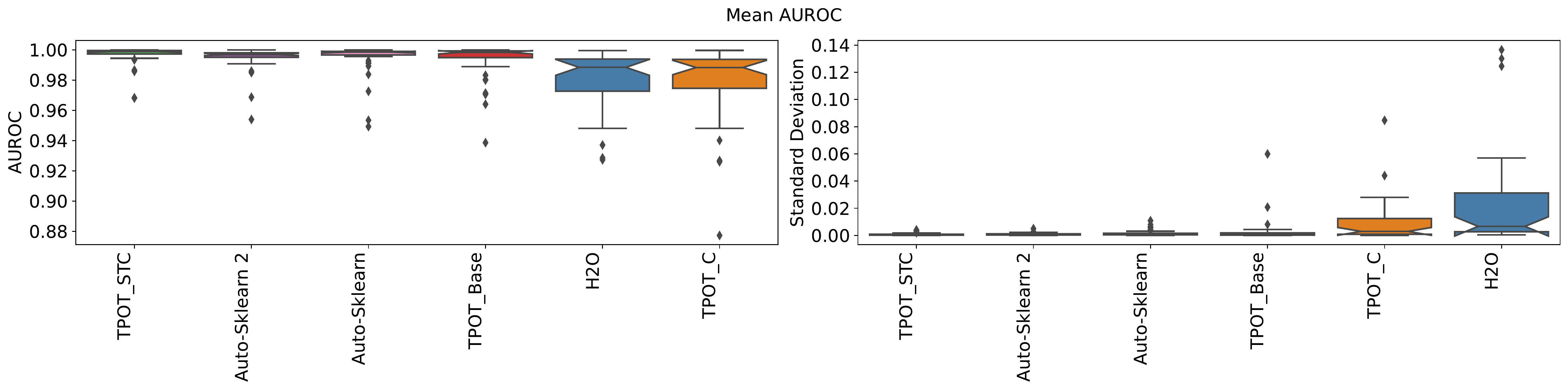}}
    \centerline{\includegraphics[scale=0.35]{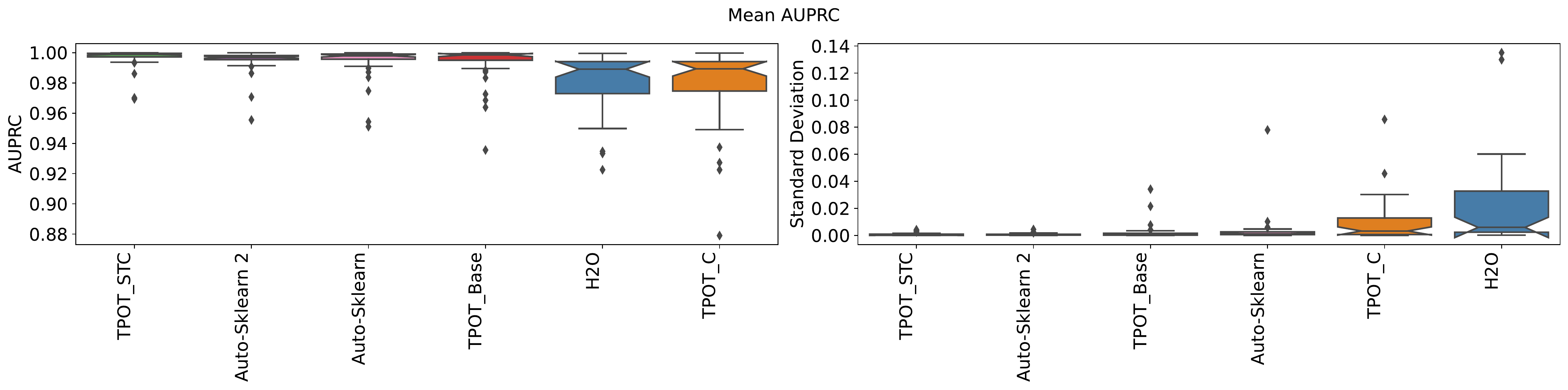}}
    \centerline{\includegraphics[scale=0.35]{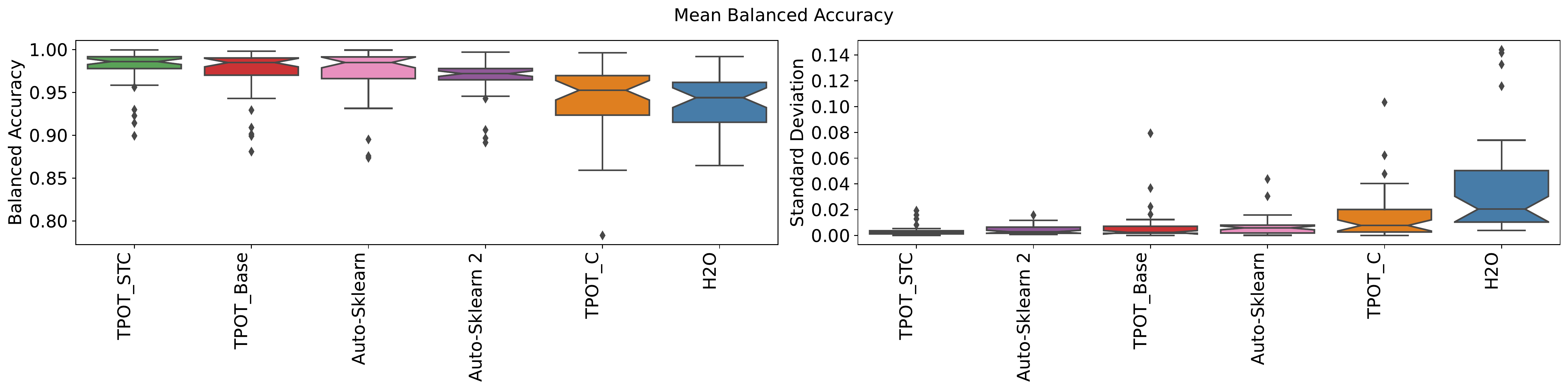}}
    \caption{First and second columns contain the distribution of average scores and standard deviations measured across all ten runs of a given dataset. }
    \label{fig:boxplots}
\end{figure}

Figure \ref{fig:strip_line} illustrates the raw results of all AutoML algorithms, datasets, and runs. Figure \ref{fig:boxplots} illustrates the average AUROC, AUPRC, and balanced accuracy for each dataset across ten runs for each of the AutoML algorithms. Additionally, we computed the standard deviation for each dataset and algorithm across the ten runs. Stochasticity in the AutoML algorithms can lead to different final models with different performances in each run. All models performed well on the datasets, though TPOT\_STC, Auto-Sklearn 1 and 2, and TPOT\_Base had the higher average scores and were more consistent run to run compared to H2O and TPOT\_C. 

\begin{figure}[htb!]
    \includegraphics[scale=0.3]{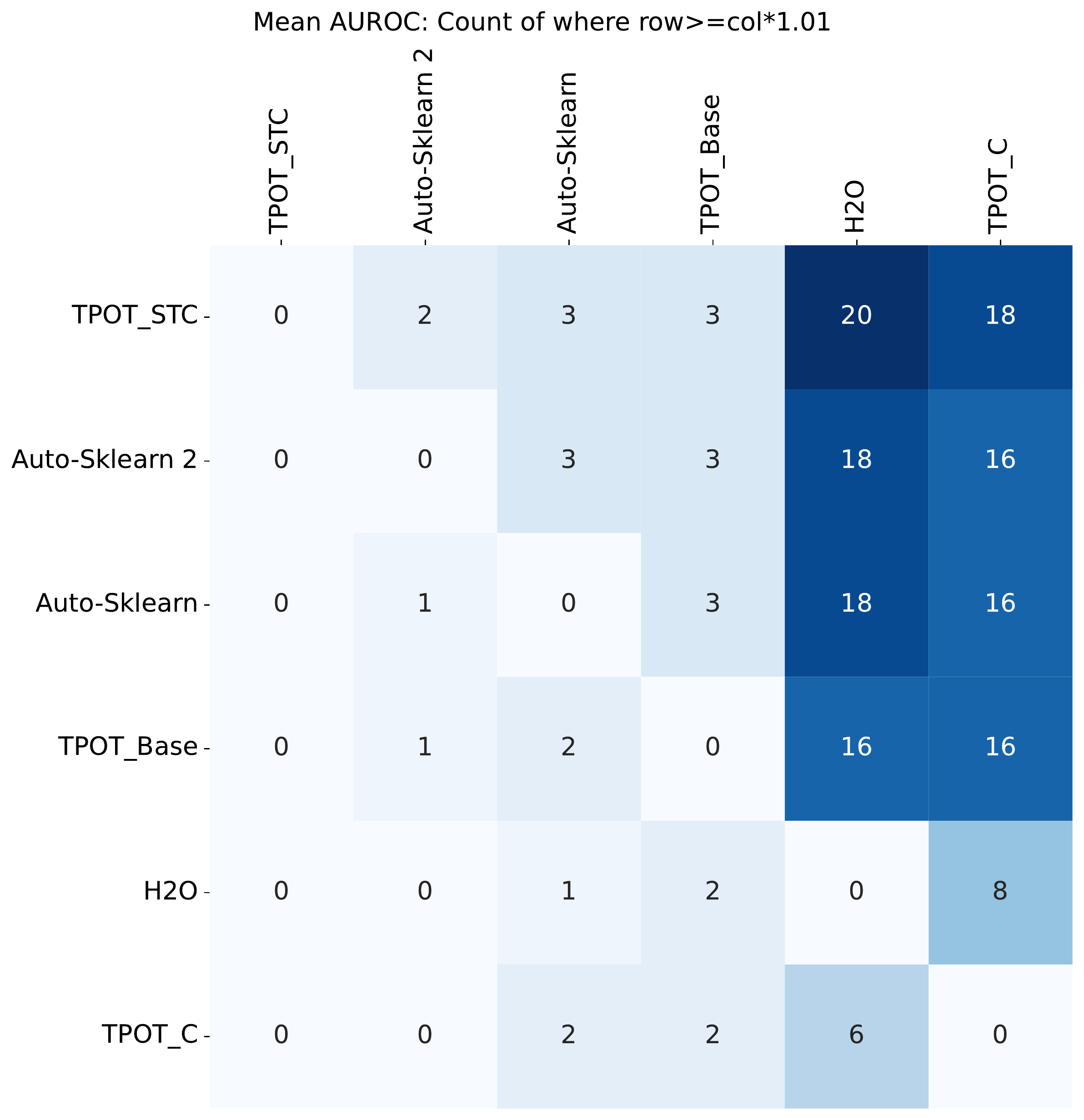}
    \includegraphics[scale=0.3]{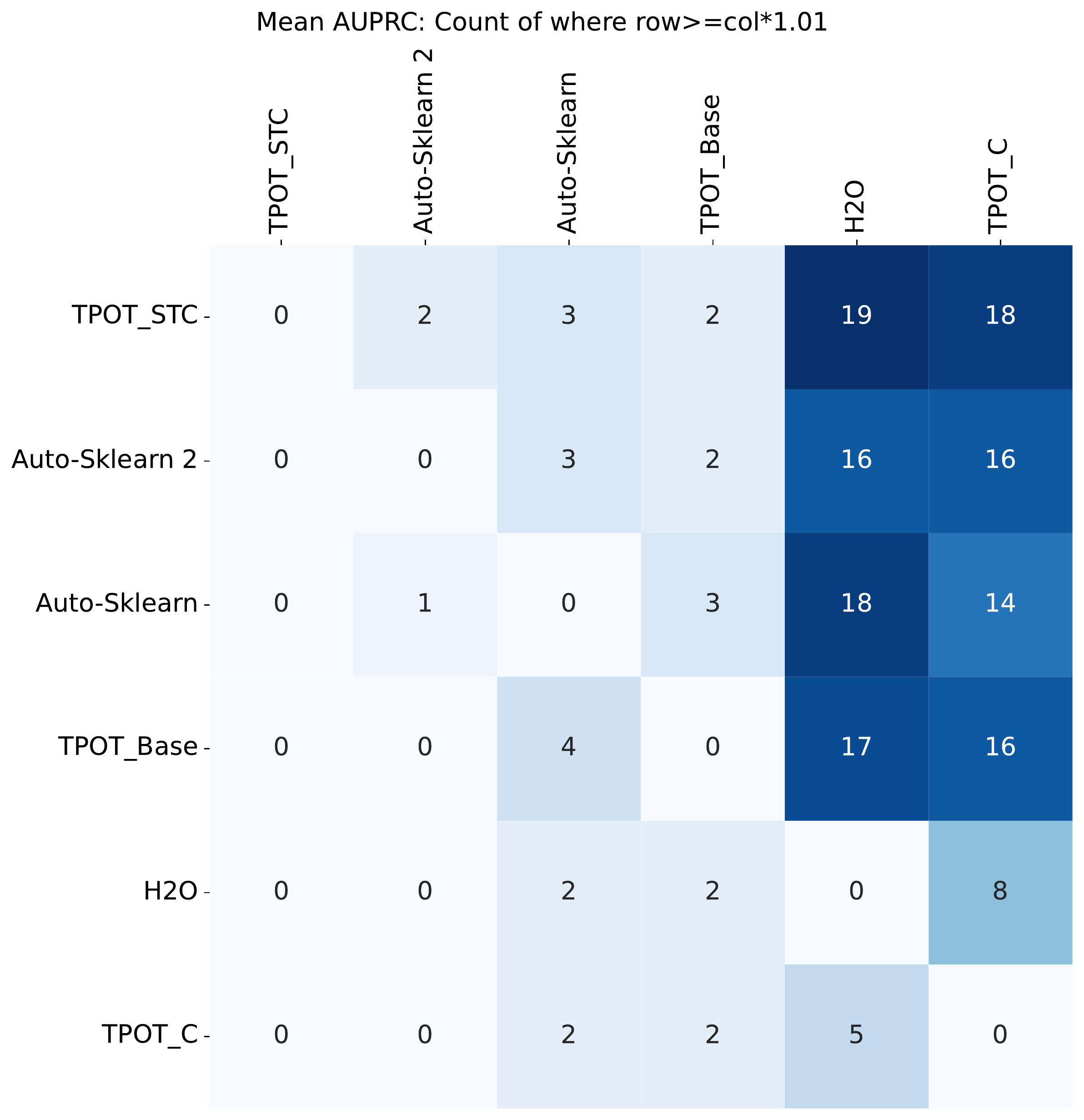}
    \includegraphics[scale=0.3]{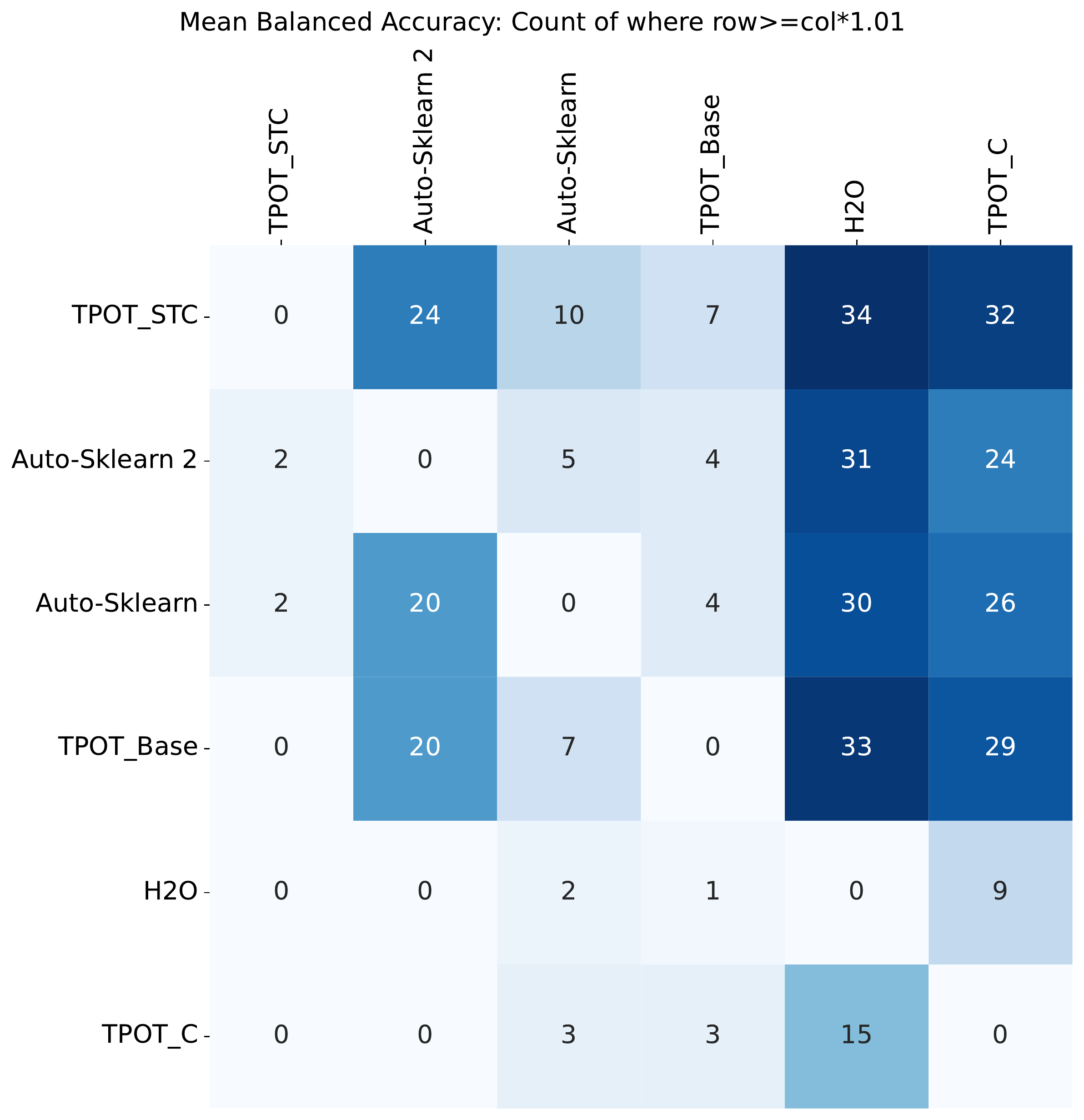}
    \caption{A matrix showing wins-loses between the pairs of the methods. Each cell represents the count of datasets where the algorithm in the row reaches at least 99\% of the score of the algorithm in the column.}
    \label{fig:wins}
\end{figure}

 Figure \ref{fig:wins} illustrates the number of times each algorithm outperforms another algorithm by 1\% of its score (i.e., the score multiplied by 1.01). For the high-performing algorithms, the performance on each dataset did not vary by a large amount. However, there were a couple of datasets where one algorithm would slightly outperform another.  

\section{Conclusion}

In this study, we demonstrated the ability of AutoML algorithms to consistently find high-performing machine-learning pipelines on various uniquely defined datasets. We also found that no one algorithm dominated all metrics on all datasets. While most algorithms performed similarly on most datasets, some outperformed others on a few datasets depending on the metric. Both TPOT (except for TPOT\_C) and Auto-Sklearn performed well on all datasets, while H2O occasionally had lower performance and suffered from consistency run-to-run.

Of the TPOT templates explored, TPOT\_STC was the most consistent run-to-run and had a slightly higher average performance. TPOT\_C was the most variable and lowest performing. The failure of TPOT\_C to find well-performing models signals that more work needs to be done to improve hyperparameter search within TPOT. Adding the feature selector and transformer to the TPOT template improved the consistency and overall performance. The same selectors and templates are found in TPOT\_Base; however, having an unrestricted pipeline slightly negatively impacted performance. One possible explanation could be that by restricting the search space, TPOT may have spent less time on unnecessarily large models. Future work on TPOT will explore better initialization and more efficient hyperparameter searching, possibly with Optuna, to improve the consistency from run to run.

As the DIGEN datasets are relatively simple and do not require heavy feature selection or transformation, it is unclear how the results from these experiments would generalize AutoML recommendations to other real-world problems. While the AutoML algorithms performed similarly in this benchmark, there may be situations where one would be preferred over others. More complex benchmarks will be required to tease apart the strengths and weaknesses of different AutoML algorithms. Datasets could also be designed to target a benchmark toward specific use cases. For example, one area of interest is heterogeneity which could be simulated by combining DIGEN equations to simulate two or more different population groups. This setup could test the ability of AutoML to simultaneously identify population groups and predict the outcome given that group. Additionally, it would be important to consider different combinations of sample size, the number of features, and generative model complexity, which could provide insight into the potential differences of AutoML algorithms under different scenarios.

As the DIGEN datasets are relatively simple and do not require heavy feature selection or transformation, it is unclear how the results from these experiments would generalize AutoML recommendations to other real-world problems. While the AutoML algorithms performed similarly in this benchmark, there may be situations where one would be preferred over others. More complex benchmarks will be required to tease apart the strengths and weaknesses of different AutoML algorithms. Datasets could also be designed to target a benchmark toward specific use cases. For example, one area of interest is heterogeneity which could be simulated by combining DIGEN equations to simulate two or more different population groups. This setup could test the ability of AutoML to simultaneously identify population groups and predict the outcome given that group. Additionally, it would be important to consider different combinations of sample size, the number of features, and generative model complexity, which could provide insight into the potential differences of AutoML algorithms under different scenarios.

Code is available in the following GitHub repository \url{https://github.com/perib/automl_digen_benchmark}

\section{Acknowledgement}

We would like to acknowledge funding from National Institutes of Health (NIH) grants U01 AG066833 and R01 LM010098.

\bibliography{mybibfile}

\end{document}